\pdfoutput=1
\documentclass[11pt, dvipsnames]{article}

\usepackage{emnlp2021}

\usepackage{times}
\usepackage{latexsym}

\usepackage[T1]{fontenc}
\usepackage[utf8]{inputenc}

\usepackage{microtype}

\usepackage{array}\newcolumntype{P}[1]{>{\centering\arraybackslash}p{#1}}
\usepackage{makecell}
\usepackage{graphicx}
\usepackage{amsmath}
\usepackage{amssymb}% http://ctan.org/pkg/amssymb
\usepackage{pifont}% http://ctan.org/pkg/pifont
\usepackage{booktabs}
\usepackage{enumitem}
\usepackage{soul}
\sethlcolor{Lavender}
\usepackage{moresize}
\usepackage{multicol}
\usepackage{multirow}
\usepackage{adjustbox}
\usepackage{subfigure}
\title{Tiered Reasoning for Intuitive Physics:\\Toward Verifiable Commonsense Language Understanding}

\author{Shane Storks$^\ast$ \hspace{20pt} Qiaozi Gao$^\dagger$  \hspace{20pt}  Yichi Zhang$^\ast$  \hspace{20pt}  Joyce
Chai$^\ast$ \\
$^\ast$Computer Science and Engineering Division, University of Michigan \\
$^\dagger$Department of Computer Science and Engineering, Michigan State University \\
  \texttt{\{sstorks, zhangyic, chaijy\}@umich.edu} \\     
  \texttt{gaoqiaoz@msu.edu}
}

\begin{document}
\maketitle
\begin{abstract}

Large-scale, pre-trained language models (LMs) have achieved human-level performance on a breadth of language understanding tasks. However, evaluations only based on end task performance shed little light on machines' true ability in language understanding and reasoning. In this paper, we highlight the importance of evaluating the underlying reasoning process in addition to end performance. Toward this goal, we introduce Tiered Reasoning for Intuitive Physics (\texttt{TRIP}), a novel commonsense reasoning dataset with dense annotations that enable multi-tiered evaluation of machines' reasoning process. Our empirical results show that while large LMs can achieve high end performance, they struggle to support their predictions with valid supporting evidence. The \texttt{TRIP} dataset and our baseline results will motivate verifiable evaluation of commonsense reasoning and facilitate future research toward developing better language understanding and reasoning models.

\end{abstract}

\section{Introduction}\label{intro}

\begin{figure*}

  \centering
  \includegraphics[width=0.96\textwidth]{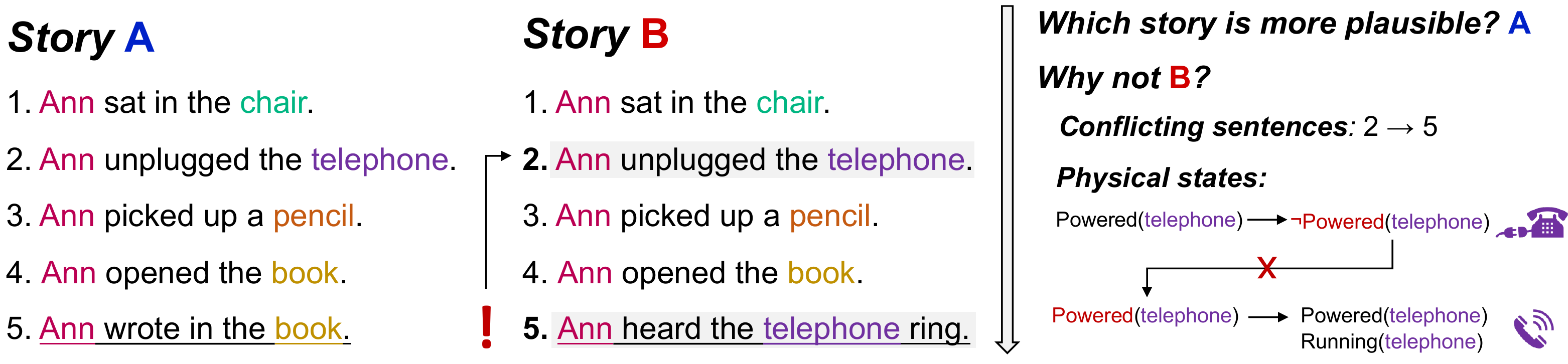}
\vspace{-6pt}

\normalsize
    \caption{Story pair from \texttt{TRIP}, along with the tiers of annotation available to represent the reasoning process.}
    \vspace{-10pt}
    \label{fig:story example}
\end{figure*}

Recent years have seen a surge of research activities toward commonsense reasoning in natural language understanding.
Dozens of relevant, large-scale benchmark datasets have been developed, and online leaderboards encourage broad participation in solving them. 
In the last few years, extraordinary performance gains on these benchmarks have come from large-scale language models (LMs) pre-trained on massive amounts of online text~\cite{petersDeepContextualizedWord2018,radfordImprovingLanguageUnderstanding2018,radfordImprovingLanguageUnderstanding2018a,raffelExploringLimitsTransfer2019,brownLanguageModelsAre2020}.
Today's best models can achieve impressive performance and have surpassed human performance in challenging language understanding tasks, including benchmarks for commonsense inference~\cite{bowmanLargeAnnotatedCorpus2015,zellersSWAGLargeScaleAdversarial2018,ch2019abductive}. This rapid period of growth and progress has been an undoubtedly exciting time for NLP.

Despite these exciting results, it is a subject of scrutiny whether these models have a deep understanding of the tasks they are applied to \cite{benderClimbingNLUMeaning2020,linzenHowCanWe2020}. 
A key concern is widespread bias in language benchmarks leading to superficial correlations between context and class labels \cite{schwartzEffectDifferentWriting2017,gururanganAnnotationArtifactsNatural2018,poliakHypothesisOnlyBaselines2018}, allowing systems to bypass reasoning and achieve artificially high performance \cite{nivenProbingNeuralNetwork2019,mccoyRightWrongReasons2019}. 
Consequently, it remains unclear whether the problems are truly solved, and whether machines can perform verifiable reasoning as humans do.

In this work, we first introduce
Tiered Reasoning for Intuitive Physics (\texttt{TRIP}), a benchmark targeting physical commonsense reasoning. \texttt{TRIP} poses a high-level end task for story plausibility classification, a common proxy task for commonsense reasoning problems \cite{roemmeleChoicePlausibleAlternatives2011,mostafazadehCorpusClozeEvaluation2016,sapSocialIQACommonsenseReasoning2019,biskPIQAReasoningPhysical2020}. Notably, however, it includes dense annotations for each story capturing multiple tiers of reasoning beyond the end task. From these annotations, we propose a tiered evaluation, where given a pair of highly similar stories (differing only by one sentence which makes one of the stories implausible), systems must jointly identify (1) the plausible story, (2) a pair of conflicting sentences in the implausible story, and (3) the underlying physical states in those sentences causing the conflict. The goal of \texttt{TRIP} is to enable a systematic evaluation of machine coherence toward the end task prediction of plausibility.
In particular, we evaluate whether a high-level plausibility prediction can be \textit{verified} based on lower-level understanding, for example, physical state changes that would support the prediction.

We further present several baseline systems powered by large LMs. Our empirical results show that while large LMs can achieve high end task performance (up to 78\% accuracy), they struggle to jointly support their predictions with the proper evidence (only up to 11\% of examples supported with correct physical states and conflicting sentences). 
Consequently, the predictions from these powerful systems are overwhelmingly not accountable to their understanding of how the world works.

The contributions of this work are the first-of-its-kind dataset \texttt{TRIP} and new metrics that facilitate quantitative evaluation of coherent reasoning in commonsense language understanding. Our detailed analysis by applying large LMs on this dataset demonstrates key disconnections between low-level and high-level predictions in the reasoning process.
This dataset and our baseline results motivate future work to develop systems that are capable of \textit{verifiable} language understanding and reasoning.

\section{Tiered Reasoning for Intuitive Physics}
Physical commonsense reasoning, also referred to as na{\"i}ve physics \cite{davisCommonsenseReasoningCommonsense2015} or intuitive physics \cite{lakeBuildingMachinesThat2017}, has recently gained attention in the NLP community \cite{gaoPhysicalCausalityAction2016,forbesVerbPhysicsRelative2017,mishraTrackingStateChanges2018,bosselutSimulatingActionDynamics2018,forbesNeuralLanguageRepresentations2019,biskPIQAReasoningPhysical2020}.
From a young age, humans possess commonsense knowledge and reasoning skills about a wide variety of physical phenomena, such as movement, rigidity, and balance \cite{blissCommonsenseReasoningPhysical2008}.
This problem is consequently thought to be especially challenging for machines because physical commonsense
is considered obvious to most humans, and suffers from reporting bias \cite{forbesVerbPhysicsRelative2017}.  
As NLP systems are typically trained only on written communications, it remains unclear whether they can learn this \cite{biskExperienceGroundsLanguage2020}.
We have developed a dataset in English to target this domain and shed more light on this question.

\subsection{\texttt{TRIP} Dataset} 
The Tiered Reasoning for Intuitive Physics (\texttt {TRIP}) is a benchmark for physical commonsense reasoning that provides traces of reasoning for an end task of plausibility prediction. 
The dataset consists of human-authored stories, such as those in Figure~\ref{fig:story example}, describing sequences of concrete physical actions. Given two stories composed of individually plausible sentences and only differing by one sentence (i.e., Sentence 5), the proposed task is to determine which story is more plausible.
To understand stories like these and make such a prediction, one must have knowledge of verb causality\footnote{For example, {\em cutting} an object causes it to be in pieces, and {\em melting} an object causes it to be in liquid form.} and precondition\footnote{For example, to {\em cut} an object, it must be in solid form, but to {\em stir} an object, it must be in liquid form.}, and 
rules of intuitive physics.\footnote{For example, the constraint that an object inside of a container moves when its container moves.} 

Plausible stories were crowd-sourced from Amazon Mechanical Turk.\footnote{\url{https://www.mturk.com/}} 
To convert each story into several implausible stories, we hired separate workers to each write a new sentence to replace a sentence in the original story, such that the new story after replacement is no longer realistic in the physical world.
To ensure quality, these workers flagged stories which were incoherent or did not describe realistic actions. We eliminated those stories and performed a manual round of validation to remove any remaining bad stories and correct typos.

\subsection{Controlled Data Curation}\label{sec:keyprops}
\texttt{TRIP} was carefully curated and restricted to support probing of reasoning abilities possessed by text classifiers. Compared to current benchmark trends, this dataset has the following unique properties.

\paragraph{Objectivity in physical commonsense.}

As commonsense knowledge differs between humans based on region, culture, and other factors~\cite{davisLogicalFormalizationsCommonsense2017}, plausible reasoning tasks can become ambiguous and subjective, for example, in open-domain commonsense reasoning problems \cite{zhangOrdinalCommonsenseInference2017,ch2019abductive}. 
To address this issue, we directed story authors to write sentences involving concrete actions, which can be unambiguously visualized in the physical world, while avoiding mental actions such as to \textit{think} or \textit{like}.
We limit stories to typical household happenings by directing annotators to write stories in one of six possible ``rooms'' seen in everyday life.

To further reduce subjectivity and block other confounding factors that may result from complex use of language, we encourage crowd workers to write sentences in a simple declarative form, typically starting with the agent of the story, followed by a verb, a direct object, and an optional indirect object. The simplicity of language use would additionally allow us to focus less on linguistic processing and semantic phenomena, and more on investigating machines' reasoning ability.

\vspace{-6pt}
\paragraph{Plausibility in longer context.}

Many benchmarks for plausible reasoning only (or most frequently) provide one sentence of context, with similarly short choices to complete the context \cite{roemmeleChoicePlausibleAlternatives2011,zellersSWAGLargeScaleAdversarial2018,biskExperienceGroundsLanguage2020}. 
In \texttt{TRIP}, we imposed several restrictions to require reasoning over multiple sentences with associated physical state changes.
First, we required annotators to write stories at least five sentences long.
Further, when collecting new sentences to convert plausible stories into implausible stories, we required that the new sentence should be plausible in isolation, and only become implausible when considering the world state implied by other sentences in the story. This constraint encourages stories to be rich in interesting action dynamics rather than nonsense sentences such as ``Mary fried eggs on the printer'' or ``Tom ate the spoon,'' which may be easier to recognize through distributional biases.
As this new sentence can conflict with any other sentence(s) in the story, solving the task requires reasoning over the entire context.

\vspace{-6pt}
\paragraph{Multi-tier annotation.}
To enable a systematic investigation of a system's reasoning process, we manually provided three levels of annotation. As shown in Figure~\ref{fig:story example}, the first level is the \textit{end task label} to indicate which of the two story choices are more plausible. 
By design, most implausible story choices have exactly one pair of \textit{conflicting sentences}, e.g., Sentences 2 and 5 in the example. 
The second level of annotation identifies these sentences in each story. 
On a random set of 100 implausible stories from the training data, a second annotator labeled these pairs of sentences, reaching a near-perfect Cohen's $\kappa$~\cite{cohenCoefficientAgreementNominal1960} of 0.929, supporting the objectivity of these labels. 
The third level justifies the implausibility with labels for the underlying \textit{physical states}, giving a detailed account of the physical changes associated with each sentence. In our example, unplugging \textit{the phone} in Sentence 2 causes it to lose power, while Sentence 5 requires that the phone is powered in order to \textit{ring}.

In order to generate these rich annotations, we defined a space of 20 physical attributes (5 for humans, 15 for objects) which capture most conflicts found in the stories. This was collected in part from related attribute spaces proposed by \citet{gaoPhysicalCausalityAction2016} and \citet{bosselutSimulatingActionDynamics2018}, and chosen based on a random set of implausible training stories, specifically the nature of their conflicts and physical changes objects underwent during the stories. For each entity in each sentence in the dataset, we annotate the implied values of these attributes before (precondition) and after (effect) the events of the sentence take place. This step of the annotation was a substantial effort. 
Note that while relevant entities in each sentence are provided in the data for convenient evaluation, these can be fairly reliably extracted using the noun chunk parser from spaCy.\footnote{\url{https://spacy.io/}}
To verify the quality of annotations, we measured inter-annotator agreement on a representative subset of 157 sentences from 31 stories in the training data, finding a substantial Cohen's $\kappa$ of 0.7917.
A detailed description of this annotation process can be found in Appendix~\ref{sec:datacollectionapx}.

\begin{table}
    \centering
    \footnotesize
    % \resizebox{.99\linewidth}{!}{
    \begin{tabular}{P{3.0cm}|P{0.6cm}P{0.6cm}P{0.6cm}|P{0.6cm}}\toprule
         \textbf{Measure} & \textbf{Train} & \textbf{Val.} & \textbf{Test} & \textbf{All} \\\midrule
         \# plausible stories & 370 & 152 & 153 & 675 \\
         \# implausible stories & 799 & 322 & 351 & 1472 \\ 
         avg. \# sentences & 5.1 & 5.0 & 5.1 & 5.1 \\
         avg. sentence length & 8.3 & 8.0 & 8.5 & 8.3 \\ \midrule
        \# story authors & 97 & 57 & 62 & 134 \\
         avg. \# stories/author & 3.8 & 2.7 & 2.5 & 5.0 \\ \midrule
         avg. \# conflicting sentence pairs & 1.2 & 1.2 & 1.2 & 1.2 \\ \midrule
         \# physical state labels & 18.8k & 8.74k & 9.09k & 36.6k \\\bottomrule
    \end{tabular}
    \normalsize
    \caption{Statistics of the \texttt{TRIP} dataset. Implausible stories in each partition are generated from and paired with the plausible stories in the same partition. }
    \vspace{-15pt}
    \label{tbl:dataset stats}
\end{table}

Table~\ref{tbl:dataset stats} lists the overall statistics of the resulting dataset. While this dataset is small by today's standards, our goal is depth, not breadth. Rather than training models on a surplus of data to simply achieve high accuracy on the end task, we aim to use our deep, multi-tiered annotations to probe the capability of NLP models to perform 
coherent reasoning toward the end task.

\subsection{Proposed Tasks}
From the \texttt{TRIP} dataset, we propose several tiered tasks as shown in Figure~\ref{fig:story example}. Together, these tasks form a human-interpretable reasoning process supported by a chain of evidence.

\paragraph{Physical state classification.}
From our physical state annotations, we propose two tasks for each sentence-entity pair in each story choice: precondition and effect state classification. For example, consider the entity \textit{potato} in the sentence ``John cut the cooked potato in half.'' First, we should predict that the potato was solid in order to be \textit{cut}, i.e., the precondition label for the \texttt{solidity} attribute is \textit{true}. Second, we should predict that the potato was in pieces as a result of being \textit{cut}, i.e., the effect label for the \texttt{in pieces} attribute is \textit{true}.

\paragraph{Conflict detection.}
Next, we define the task of conflict detection as identifying a pair of sentences in the form $S_i \rightarrow S_j$. $S_j$ is a \textit{breakpoint}, i.e., the point where the story first becomes implausible given the context so far, while $S_i$ serves as \textit{evidence} that explains the breakpoint, usually causing a conflicting world state. For example, in Figure~\ref{fig:story example}, Sentence 5 is a breakpoint, while Sentence 2 is the evidence that explains why the story becomes implausible after Sentence 5. Note that it is possible that a story may have multiple pairs of conflicting sentences beyond the breakpoint and evidence pair. However, across the dataset, the average number of conflicting sentence pairs is only 1.2, so one conflicting sentence pair is a sufficient and simpler explanation for the conflict (albeit not exhaustive).

\paragraph{Story classification.}
Lastly, the end task is to determine which of two stories %, %just one of which is implausible, 
is the plausible one. This should be determined based on any conflicts detected within the two stories.

\vspace{5pt}

\subsection{Benchmark Goals}

It is important to note that while one can treat these tasks separately, the goal of this benchmark is to solve them jointly to form a coherent reasoning chain: physical state classification explains conflict detection, which further explains story classification. 
Unlike most existing benchmarks in this area, which assess language understanding ability through some high-level end tasks, the goal of our benchmark is to enable development of systems for interpretable and consistent reasoning toward language understanding.
Our baseline models (Section~\ref{sec:tieredbaseline}) and evaluation metrics (Section~\ref{sec:metrics}) are developed to serve this purpose.

It is also worth noting that although data bias is an issue for high-level benchmark tasks where systems are not required to justify their predictions, we are not directly targeting this issue. Recent work has attempted to remove biases from benchmark data and thus prevent exploitation of them in performing high-level tasks \cite{zellersSWAGLargeScaleAdversarial2018,nieAdversarialNLINew2020}. In contrast, our framing of language understanding as being built from the ground up (i.e., from low-level to high-level tasks) provides systems with the proper supporting evidence toward high-level tasks, and thus can potentially mitigate some of the problems around data bias. 

\begin{figure*}
    \centering
    \includegraphics[width=0.99\textwidth]{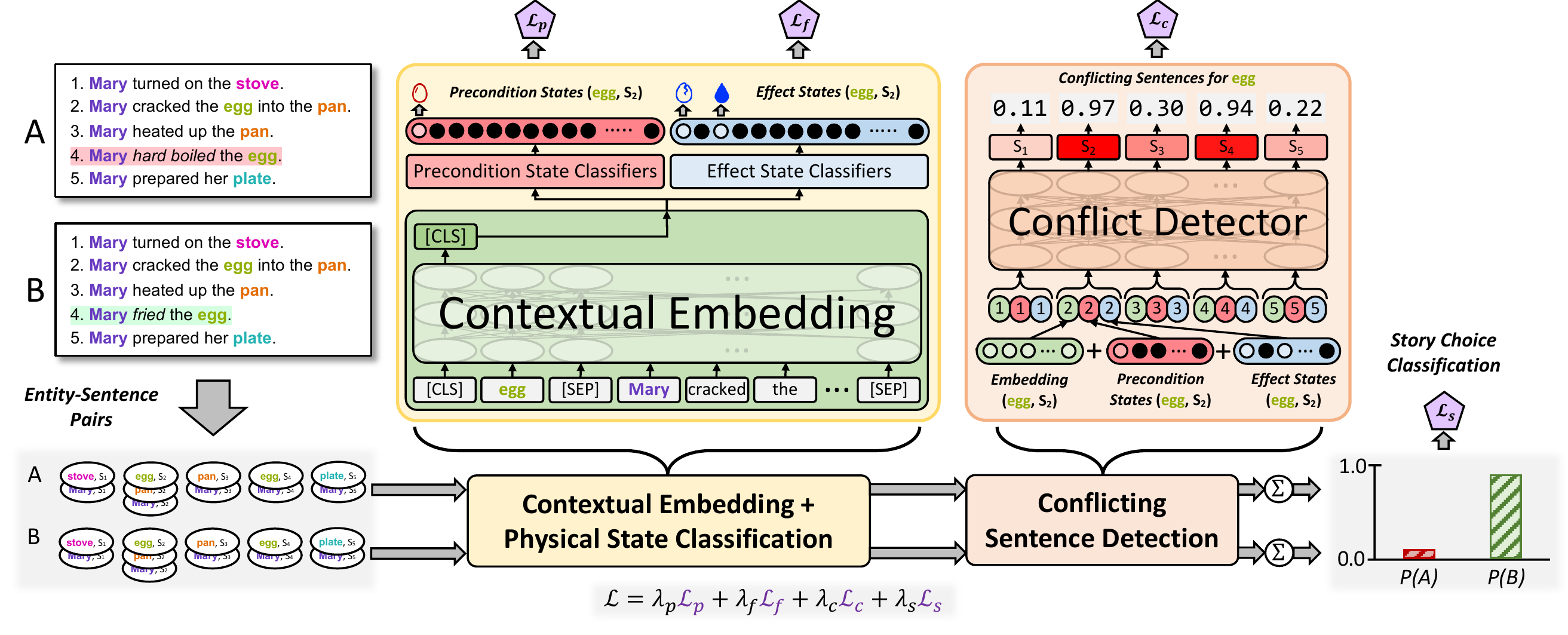}
    \vspace{-10pt}
    \caption{Proposed tiered reasoning system with loss functions $\mathcal{L}_p$ for precondition state classification, $\mathcal{L}_f$ for effect state classification, $\mathcal{L}_c$ for conflicting sentence detection, and $\mathcal{L}_s$ for story choice classification. The model is trained end-to-end by optimizing the joint loss $\mathcal{L}$, a weighted sum of these loss functions.}
    \vspace{-10pt}
    \label{fig: tiered prototype}
\end{figure*}

\section{A Tiered Baseline for \texttt{TRIP}}
\label{sec:tieredbaseline}

Figure~\ref{fig: tiered prototype} displays a high-level view of our proposed baseline system to solve \texttt{TRIP}. It individually embeds each sentence-entity pair in each story, classifies physical precondition and effect states, then identifies conflicting sentences from these. Given a pair of stories, it aggregates conflict predictions for each story to decide which is more plausible.

\subsection{Module Implementations}
Each module is implemented as some kind of neural network architecture. Here, we describe some details of the implementations.

\paragraph{Contextual Embedding.}
The Contextual Embedding module is implemented as a pre-trained, transformer-based language model. Generally, this module takes as input a sentence and the name of an entity from a story, following an entity-first input formulation \cite{guptaEffectiveUseTransformer2019}, and outputs a dense, contextualized numerical representation.

\paragraph{Precondition and Effect Classifiers.}
The Precondition and Effect Classifiers are implemented as typical feedforward classification heads for contextual embeddings, with one precondition classifier and one effect classifier for each of the 20 physical attribute tracked in the dataset.
Softmax is applied to the output for classification. Altogether, the predictions from these classifiers label physical states of each entity in each sentence of the story.

\paragraph{Conflict Detector.}
For each entity and its predicted physical states over all sentences in a story, the Conflict Detector predicts whether there is some conflict in the entity's physical states, specifically flagging a pair of conflicting sentences through multi-label classification. We use another transformer for this module, but model the high-level sequence of sentences in a story rather than the low-level sequence of tokens in a sentence.
For each sentence-entity pair, we input the contextual embedding, as well as the classification logits behind all physical state predictions. 
We apply an additional feedforward classification layer and sigmoid function to the generated hidden states in order to model the belief probability of each sentence conflicting with another sentence in the story.

\paragraph{Story choice prediction.}
Given any detected conflicts, we lastly select which of the two given stories is plausible. As each Conflict Detector output represents a belief that the physical states of an entity in a particular sentence conflict with that of another sentence, we can simply sum the negative outputs for each story and apply softmax to determine which story is least likely to have a conflict.

\subsection{Model Training}
We train the architecture's parameters through gradient descent on the overall loss $\mathcal{L}$:

\vspace{-20pt}
\begin{gather*}
    \mathcal{L} = \lambda_p \mathcal{L}_p + \lambda_f \mathcal{L}_f + \lambda_c \mathcal{L}_c + \lambda_s \mathcal{L}_s
\end{gather*}
\vspace{-20pt}

$\mathcal{L}$ sums individual cross-entropy loss functions $\mathcal{L}_p$ for precondition classification, $\mathcal{L}_f$ for effect classification, $\mathcal{L}_c$ for conflict detection, and $\mathcal{L}_s$ for story choice classification, each balanced by respective weights $\lambda_p$, $\lambda_f$, $\lambda_c$, $\lambda_s$ summing to 1. 

\section{Experiments}\label{sec:results}
\label{sec:evaluation}

Using \texttt{TRIP}, we evaluate several variations of the proposed reasoning system powered by selected pre-trained language models: \textsc{BERT}~\cite{devlinBERTPretrainingDeep2018}, \textsc{RoBERTa}~\cite{liuRoBERTaRobustlyOptimized2019}, and \textsc{DeBERTa}~\cite{heDeBERTaDecodingenhancedBERT2021}.\footnote{We use the ``large'' configurations of \textsc{BERT} (355M parameters) and \textsc{RoBERTa} (355M parameters), and the ``base'' configuration of \textsc{DeBERTa} (140M parameters).} 
These models offer a range in design choices such as model complexity and size of pre-training data.
We begin with an evaluation from the perspective of the end task, then take a detailed look at the lower-level tasks.

\subsection{Evaluation Metrics}\label{sec:metrics}

To enable a better understanding of machines' ability in coherent reasoning toward end task performance, we apply the following evaluation metrics.

\vspace{-5pt}
\paragraph{Accuracy.} The traditional metric of end task accuracy, i.e., the proportion of testing examples where plausible stories are correctly identified. 

\vspace{-5pt}
\paragraph{Consistency.} The proportion of testing examples where not only the plausible story is correctly identified, but also the conflicting sentence pair for the implausible story is correctly identified. This is to demonstrate the consistency with identified conflicts when reasoning about plausibility.

\vspace{-5pt}
\paragraph{Verifiability.} The proportion of testing examples where not only the plausible story and the conflicting sentence pair for the implausible story are correctly identified, but also underlying physical states (i.e., preconditions and effects) that contribute to the conflict are correctly identified.\footnote{At least one nontrivial, i.e., non-default, positive-class physical state label must be predicted in the preconditions of the breakpoint sentence and effects of the evidence sentence, and all such predictions must be correct.} This is to demonstrate that the detected conflict can be verified by a correct understanding of the underlying implausible change of physical states.

It is worth noting that this notion of verifiability, although different, is motivated by the notion of \textit{verification} in software engineering \cite{verification}. This term refers to determining whether a given software solution satisfies its architectural and design requirements, and is built from the correct sub-components. Along this line, our notion of verifiability can be seen as a method to evaluate whether a language understanding system's reasoning process is built up from the correct components.

Each successive metric dives deeper into the coherence of reasoning that supports the end task prediction. Consequently, if accuracy is $a$, consistency is $b$, and verifiability is $c$, then $a \geq b \geq c$. A system that reliably produces a coherent chain of reasoning is demonstrated by $a \approx b \approx c$.

\subsection{Results}\label{sec:sub results}

Recall that we consider four loss functions for training the tiered system: $\mathcal{L}_p$ for precondition classification, $\mathcal{L}_f$ for effect classification, $\mathcal{L}_c$ for conflicting sentence detection, and $\mathcal{L}_s$ for story choice classification. To investigate how each loss affects model performance, we train instances using several combinations of them. The results of this study on the validation set are listed in Table~\ref{tbl:loss results}.

\begin{table}
    \centering
    \footnotesize

    \begin{tabular}{P{1.4cm}|P{1.3cm}P{1.6cm}P{1.6cm}}\toprule
             & \textbf{Accuracy} & \textbf{Consistency} & \textbf{Verifiability} \\
            \textbf{Model}  &   (\%) & (\%) & (\%)    \\\midrule
        {random} &  47.8 &  11.3  & 0.0   \\\midrule
        \multicolumn{4}{c}{\textit{All Losses}}  \\\midrule                    
        \textsc{BERT} & \textbf{78.3} & 2.8 & 0.0   \\ 
        \textsc{RoBERTa} & 75.2 & 6.8 & 0.9    \\
        \textsc{DeBERTa}  & 74.8 & 2.2 & 0.0    \\\midrule
        \multicolumn{4}{c}{\textit{Omit Story Choice Loss} $\mathcal{L}_s$}  \\\midrule        
        \textsc{BERT}  & 73.9 & \textbf{28.0} & 9.0     \\
       \textsc{RoBERTa}   & 73.6 & 22.4 & \textbf{10.6} \\ 
        \textsc{DeBERTa} & 75.8 & 24.8 & 7.5   \\\midrule
        \multicolumn{4}{c}{\textit{Omit Conflict Detection Loss} $\mathcal{L}_c$}  \\\midrule        
        \textsc{BERT} & 50.9 & 0.0 & 0.0   \\
        \textsc{RoBERTa} & 49.7 & 0.0 & 0.0  \\
        \textsc{DeBERTa} & 52.2 & 0.0 & 0.0     \\\midrule
        \multicolumn{4}{c}{\textit{Omit State Classification Losses} $\mathcal{L}_p$ and $\mathcal{L}_f$} \\\midrule                
        \textsc{BERT} & 75.2 & 17.4 & 0.0    \\
        \textsc{RoBERTa}  & 71.4 & 2.5 & 0.0   \\
        \textsc{DeBERTa}  & 72.4 & 9.6 & 0.0 \\\bottomrule
    \end{tabular}

    \normalsize
    \caption{End and tiered task metrics for tiered classifiers on the validation set of \texttt{TRIP} trained on varied combinations of loss functions. Random baseline (averaged over 10 runs) makes tiered predictions at random.}
    \vspace{-10pt}
    \label{tbl:loss results}
\end{table}

\paragraph{The role of end task supervision.}
In the first section of Table~\ref{tbl:loss results}, we train the system jointly on all four loss functions. Here, we see low verifiability and consistency for all three LMs, while the end task accuracy is relatively high, reaching 78.3\% when using \textsc{BERT}. When we omit the story classification loss in the second section, however, we see sharp gains in verifiability and consistency for all models, with \textsc{RoBERTa} jumping from 0.9\% verifiability and 6.8\% consistency to 10.6\% and 22.4\%, respectively. This comes at a slight cost of end task accuracy for \textsc{BERT} and \textsc{RoBERTa}. 

This suggests that while fine-tuning systems based on a high-level classification loss targeting the end task can improve the end task accuracy, this drastically reduces the interpretability of the underlying reasoning process. One potential explanation for this is that this loss drives the system to exploit spurious statistical cues in order to further increase the end task accuracy.
This gives us motivation to move away from using over-simplified end tasks to train and evaluate language understanding. In fact, if we fine-tune \textsc{RoBERTa}'s contextual embedding directly on the end task of \texttt{TRIP} without intermediate classification layers, we can achieve up to 97\% accuracy, but have no insight toward verifiability or consistency of the system. This raises questions about the validity of such a result.

\paragraph{Natural emergence of intermediate predictions.}
In the third and fourth sections of Table~\ref{tbl:loss results}, we respectively omit conflict detection loss and state classification losses to explore whether conflicting sentences or physical states would emerge naturally in the reasoning process. When omitting conflict detection loss, all metrics degrade to near or below random performance. Clearly, conflict detection is not implicitly learned from the downstream story classification loss, and since the story choice classification directly depends on the conflict detection output, the end task accuracy drops as well. 

Meanwhile, when omitting physical state classification loss, verifiability unsurprisingly drops to zero, but high accuracy on the end task can still be achieved by all models (up to 75.2\%). Notably, this suggests that reasonable supporting evidence is not required in order to achieve high accuracy on the end task.
This casts further doubt that existing state-of-the-art results on other commonsense language understanding benchmarks possess any kind of coherent reasoning beyond end classification tasks which over-simplify the problem.

In Table~\ref{tbl: test results}, we present the testing results for the best loss function configuration of the system, i.e., omitting story choice classification loss. Compared to the validation set results in Table~\ref{tbl:loss results}, we see slight drops in consistency and verifiability, further demonstrating the difficulty of this problem.

\begin{table}
    \centering
    \footnotesize
    
    \begin{tabular}{P{1.4cm}|P{1.3cm}P{1.6cm}P{1.6cm}}\toprule
            & \textbf{Accuracy} & \textbf{Consistency} & \textbf{Verifiability} \\
            \textbf{Model} &  (\%) & (\%)  &   (\%)    \\\midrule
        {random} &  49.5 &  10.7  & 0.0\\\midrule

        \textsc{BERT} & 70.9  & 21.9 & 8.3  \\
        \textsc{RoBERTa}  &  \textbf{72.9} & 19.1 & \textbf{9.1}  \\
        \textsc{DeBERTa}  & \textbf{72.9}  & \textbf{22.2} & 6.6    \\\bottomrule

    \end{tabular}

    \normalsize
    \caption{Metrics for the best tiered systems on the test set of \texttt{TRIP}. Compared to random baseline.}
    \vspace{-10pt}
    \label{tbl: test results}
\end{table}

\subsection{Analysis}
Given the poor performance along our proposed metrics, we next consider the connections between the tiered tasks, and what goes wrong in unverifiable end task instances. We focus our analysis on the systems achieving the highest verifiability on the validation set in Section~\ref{sec:sub results}.

\begin{figure}
    \centering
    \includegraphics[width=0.9\linewidth]{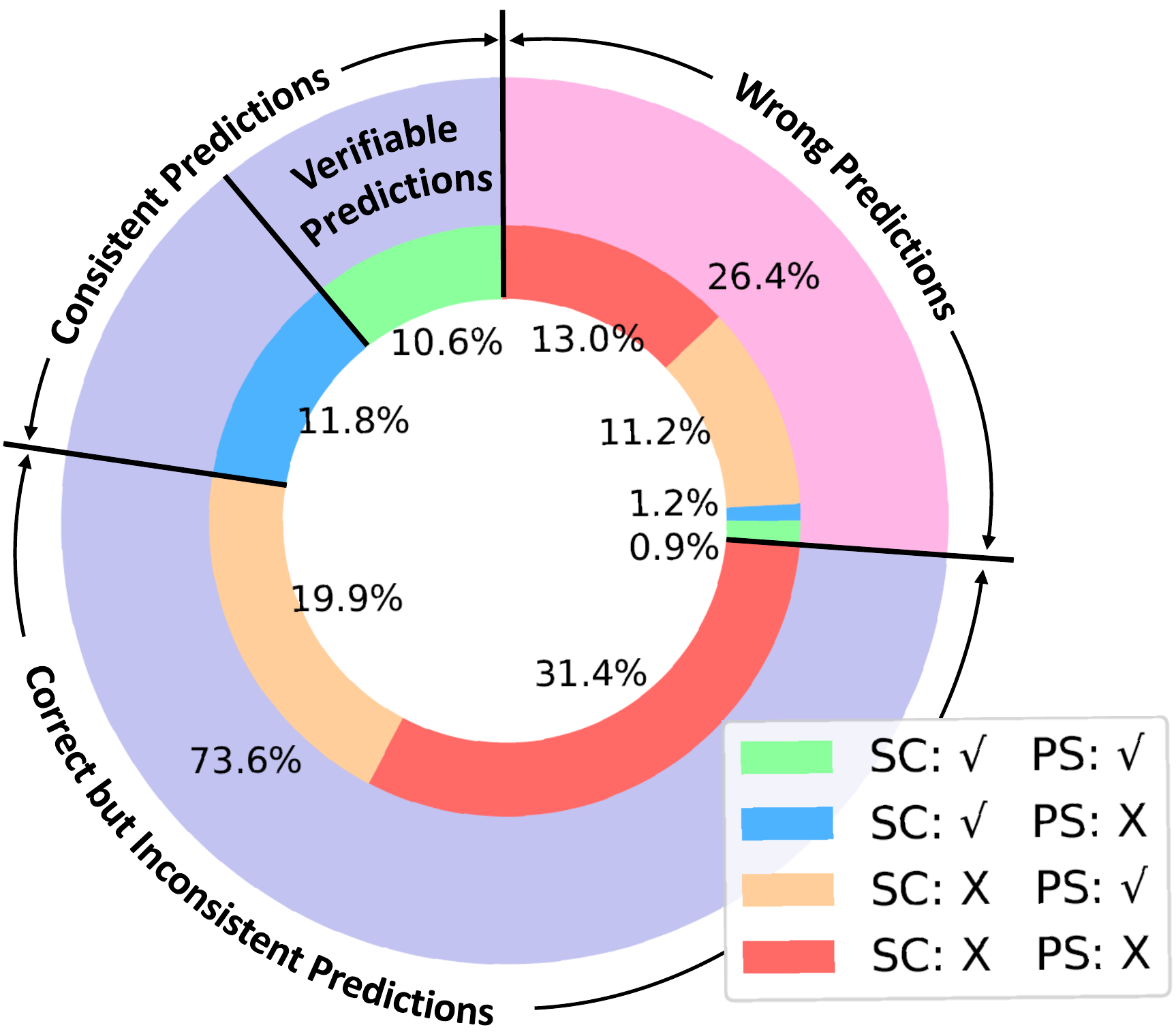}
    \caption{Distribution of \textsc{RoBERTa} successes and failures on \texttt{TRIP}. SC (sentence conflict) and PS (physical state) denote whether the predicted conflicting sentences or physical states are correct ($\checkmark$) or not ($\times$).}
    \label{fig:error_analysis}
\end{figure}

\paragraph{Failure mode distribution.}
Figure~\ref{fig:error_analysis} provides a detailed breakdown of the combinations of failure modes on the validation set. Of the 73.6\% of validation instances that are classified correctly on the end task, almost half of these (31.4\% overall) are entirely unverified, with incorrect physical states and conflicts predicted by the system. Similarly, of the 26.4\% of instances with \textit{incorrect} end task predictions, about half (13\% overall) have incorrect physical state and conflict predictions.
Meanwhile, a combined 31.1\% of instances correctly predict physical states in the conflicting sentences of the implausible story, but fail to detect a conflict in those sentences (19.9\% are correct at the end task, while 11.2\% are not). These instances, represented by orange wedges in the graph, are a significant disconnect in the reasoning process.

\begin{table}
    \centering
    \footnotesize
    
    \begin{tabular}{P{1.4cm}|P{1.5cm}P{1.5cm}P{1.5cm}}\toprule
            &  \textbf{Prec. F1} & \textbf{Eff. F1}  & \textbf{Confl. F1} \\
            \textbf{Model} &    (\%) & (\%) & (\%)   \\\midrule

        \textsc{BERT} & \textbf{54.9} & 57.2 & 66.3    \\
       \textsc{RoBERTa}  &  51.2 & 51.2 & \textbf{69.6} \\
        \textsc{DeBERTa}  & 52.8 & \textbf{57.3} & 63.6 \\\bottomrule      
        
    \end{tabular}

    \normalsize
    \caption{Macro-F1 scores of best tiered systems on aggregate precondition, effect, and conflicting sentence classification. Scores averaged over all attributes for physical state classification.}
    \label{tbl: agg measures}
\end{table}

\paragraph{Low-level task performance.}
To further address this disconnect, 
we examined system performance from the perspective of physical state classification and conflict detection. First, Table~\ref{tbl: agg measures} lists the validation metrics for our best baselines on the tasks of precondition and effect classification (by sentence-entity pair), as well as conflicting sentence detection (by end task instance). Across the board, %we find fairly high performance on all tasks.
we find reasonable performance on all tasks.

\begin{figure*}
    \centering
    \includegraphics[width=0.98\linewidth]{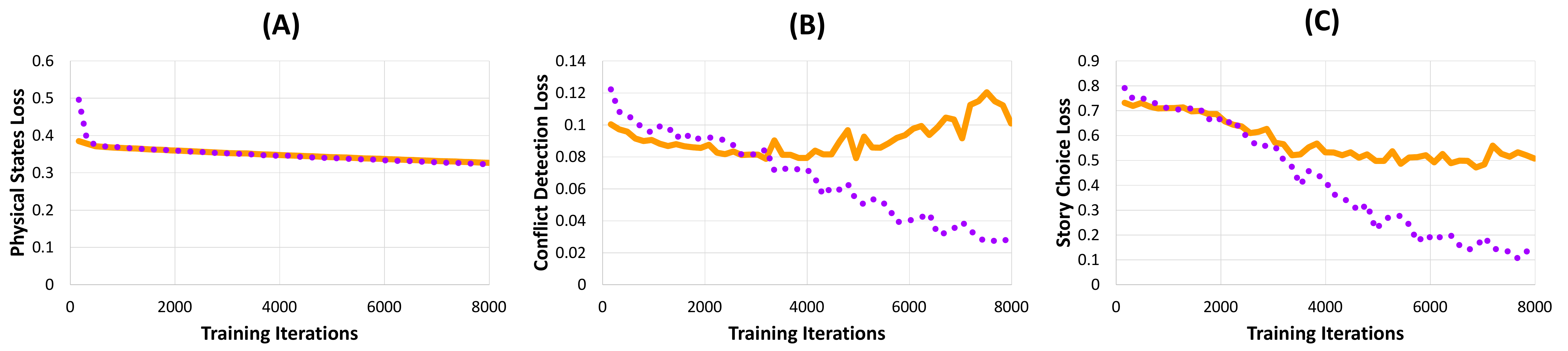}
    \caption{Training (purple, dotted) and validation (orange, solid) losses for best tiered \textsc{RoBERTa} system trained on \texttt{TRIP} for 10 epochs. Uses the best configuration of the loss functions (as found in Section~\ref{sec:sub results}) for (A) physical state classification, (B) conflict detection, and (C) story choice classification. Validation loss recorded 4 times per epoch, with training loss averaged over the trained batches since the previous recording.}
   \vspace{-10pt}
    \label{fig: learning curves}
\end{figure*}

The best performing baseline from Table~\ref{tbl:loss results} is trained using loss functions for both physical state classification and conflict detection. Given this configuration, we further examined how each task is learned.
Figure~\ref{fig: learning curves} shows training curves for the loss functions of physical state classification (averaged for precondition and effect), conflicting sentence detection, and story choice classification. Notably, though story choice classification is not used as a training objective, this end task is learned fairly well (albeit overfitting), with training and validation losses generally decreasing through training. This shows that learning to reason from the lower-level tasks is successful to some degree. However, the lower-level tasks appear challenging to learn. For physical state classification, losses decrease steadily, but slowly. For conflict detection, the losses also decrease slowly, and the model begins overfitting the training data, perhaps indicating a need for more training data at this challenging step. Future work may consider automatic data augmentation techniques to resolve this.

\begin{figure}
    \centering
    \includegraphics[width=0.98\linewidth]{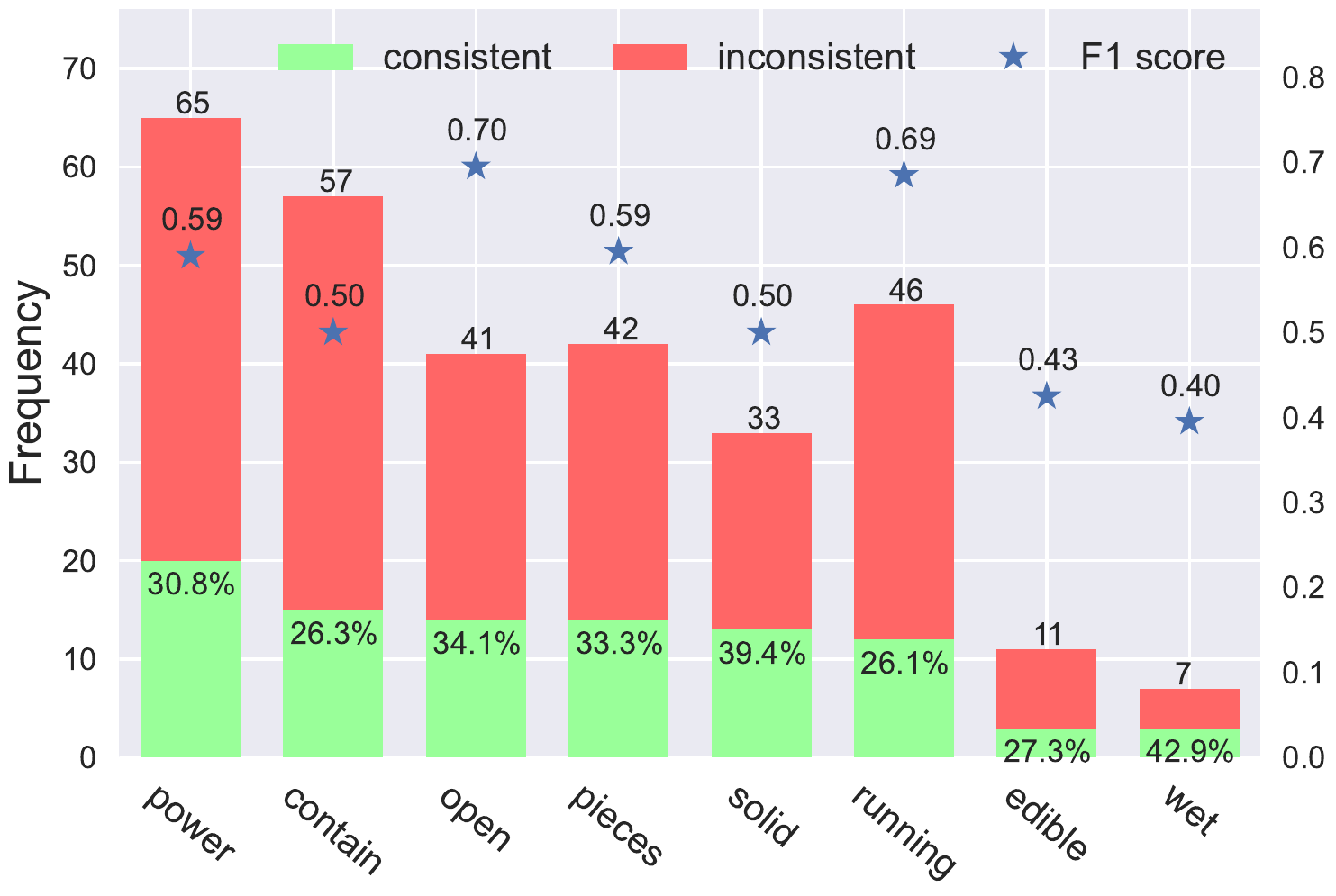}
    \vspace{-10pt}
    \caption{Contribution of correct \textsc{RoBERTa}-predicted physical states to consistency evaluation for selected attributes. The macro-F1 score of precondition and effect predictions is shown by blue stars. Among all correctly predicted states (for both effects and preconditions), the bar regions indicate whether these states appear in successfully detected conflicting sentences. }
    \vspace{-10pt}
    \label{fig:attr-to-conflict}
\end{figure}

\paragraph{Connecting states to conflicts.}
To dig deeper into the connection between physical states and plausibility conflicts, we next examined correct physical state predictions by attribute in Figure~\ref{fig:attr-to-conflict}. In the graph, we indicate the percentage of predictions supporting a successfully detected conflict, which may be interpreted as a \textit{utility} measure of each attribute toward conflict detection. We find that some attributes, like whether an electrical object is \texttt{running}, rarely contribute to successful conflict detections (only 26.1\%) despite having reasonably high F1 score (0.69). Other attributes, like \texttt{wet}, are more likely to appear in successful conflict detections when predicted correctly, even though their overall classification performance is lower. 
This provides strong insights for targeted improvement, for example, to better take advantage of lower-level predictions toward high-level tasks.

\begin{figure}[t]
	\centering
		\subfigure[A verifiable prediction.]
		{	\label{fig:case_verifiable}
			\includegraphics[width=0.96\columnwidth]{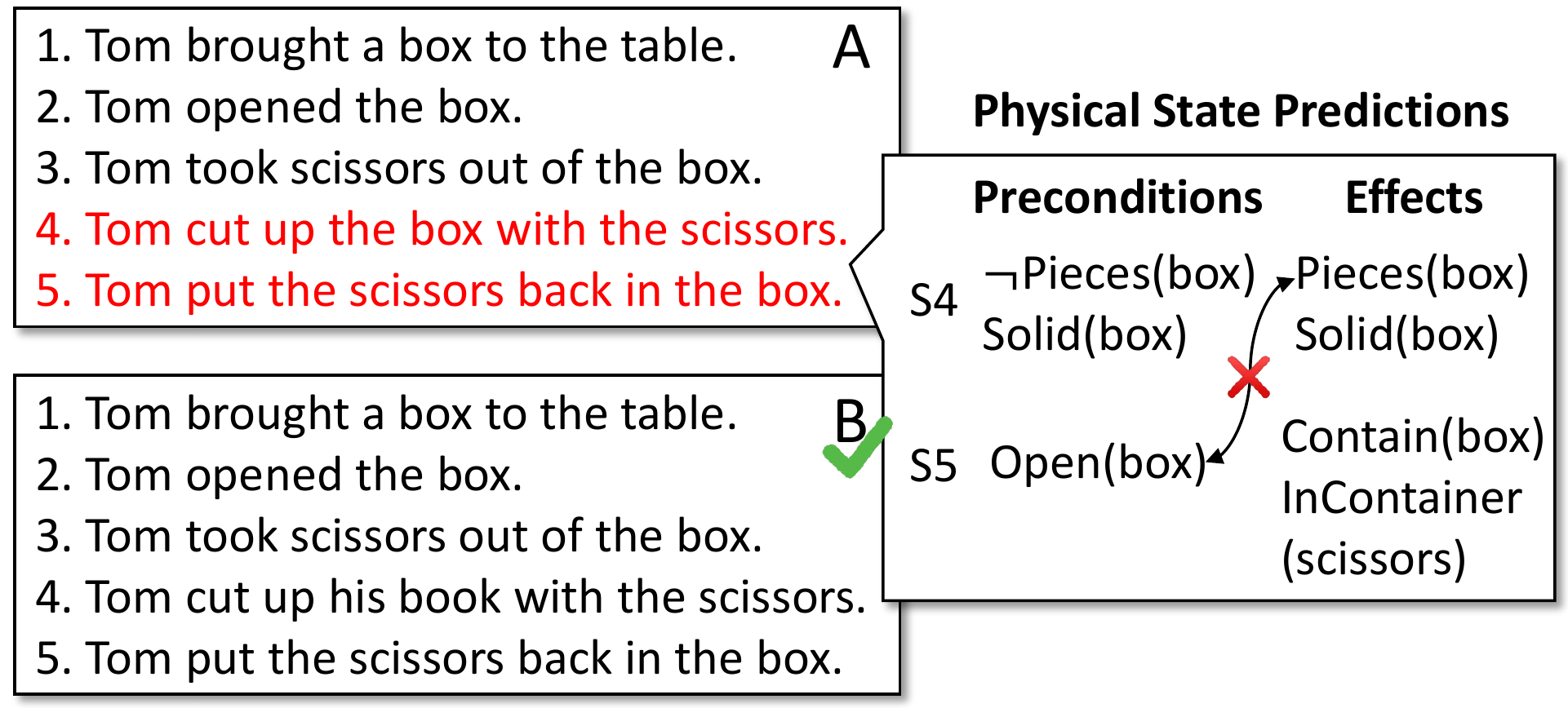} }
		\subfigure[A consistent but not verifiable prediction.]
		{	\label{fig:case_consistent}
			\includegraphics[width=0.96\columnwidth]{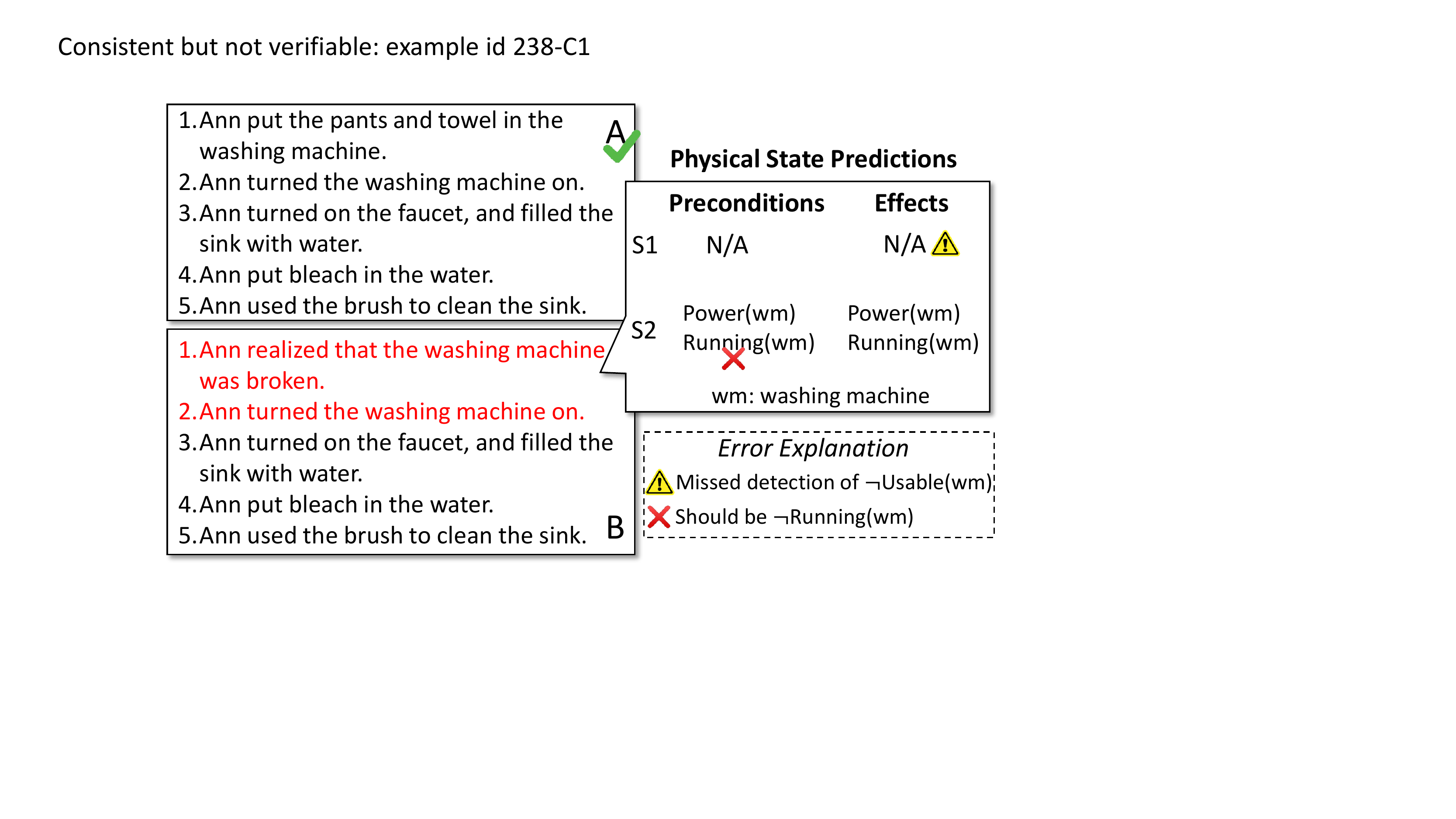} }
    \vspace{-10pt}			
	\caption{Sample outputs from the baseline system. The detected conflicting sentences are in red, and physical state predictions are shown on the right.}
	\vspace{-15pt}			
	\label{fig:case_study}
\end{figure}

\paragraph{Sample system outputs.}
Figure~\ref{fig:case_study} presents sample outputs from the tiered \textsc{RoBERTa} system. In Example (a), the prediction is entirely verifiable. The system correctly chooses the plausible story, identifies Sentences 4 and 5 as the conflicting sentences in the implausible story, and even predicts that the \textit{box} is \texttt{in pieces} after Sentence 4, and thus cannot become \texttt{open} in Sentence 5. In Example (b), the prediction is consistent but unverifiable, as the system identifies a conflict between Sentences 1 and 2, but cannot support the conflict with correct underlying physical states in either sentence. Although some relevant attributes are identified for the breakpoint sentence, e.g., \texttt{power} and \texttt{running}, they are not quite right. Meanwhile, no states are predicted for the evidence sentence.

\section{Related Work}

\paragraph{Physical commonsense.}
There exist a few NLP datasets around physical commonsense reasoning which offer various classification tasks.
ProPara~\cite{mishraTrackingStateChanges2018} tracks existence and location of entities in each sentence, similar to \texttt{TRIP}'s physical state classification, but in a more restricted state space. Physical Interaction Question Answering (PIQA) from \citet{biskPIQAReasoningPhysical2020} provides a similar high-level end task of multiple-choice text plausibility classification targeting physical commonsense.
Other benchmarks focus on specific domains of physical reasoning, such as temporal reasoning~\cite{zhou2019going} and spatial reasoning~\cite{mirzaeeSpartQATextualQuestion2021}. Visual~\cite{johnsonCLEVRDiagnosticDataset2017,bakhtin2019phyre} and multimodal~\cite{hudson2018gqa,dasEmbodiedQuestionAnswering2018,andersonVisionandLanguageNavigationInterpreting2018,ALFRED20} benchmarks also investigate systems' commonsense understanding of the physical world through perception and interaction.
Different from these existing benchmarks, \texttt{TRIP} is the first dataset of its kind with dense annotation to support evaluation of verifiable reasoning toward the end task prediction.

\paragraph{Robust language inference.}
In the face of statistical bias enabling artificially high performance in NLP models, several works have explored ways to evaluate and enable robust language inference.
Several probing studies have examined how well surface-level syntactic and semantic phenomena are captured in contextual language embeddings \cite{adiFinegrainedAnalysisSentence2017,ettingerAssessingCompositionSentence2018,tenneyWhatYouLearn2018,hewittStructuralProbeFinding2019,jawaharWhatDoesBERT2019,tenneyBERTRediscoversClassical2019a}. For stronger evaluation of potentially biased systems, others have explored specialized natural language inference tasks~\cite{welleck-etal-2019-dialogue,uppal-etal-2020-two} and logic rules~\cite{li-etal-2019-logic,asai-hajishirzi-2020-logic} to support and evaluate consistency of models across instances of the end task.
Some approaches have been proposed to instead remove biases from language by filtering out data too easily discriminated by state-of-the-art text classifiers \cite{zellersSWAGLargeScaleAdversarial2018,nieAdversarialNLINew2020}, and to improve robustness of systems against exploiting various types of biases \cite{belinkovDonTakePremise2019,clarkDonTakeEasy2019,minSyntacticDataAugmentation2020}.
Recent work has attempted to compile large amounts of semi-structured commonsense knowledge \cite{sapATOMICAtlasMachine2019,mostafazadehGLUCOSEGeneraLizedCOntextualized2020} and inject this knowledge into pre-trained language models \cite{bosselutCOMETCommonsenseTransformers2019,zhangERNIEEnhancedLanguage2019} in order to enable knowledge-supported language understanding and on-the-fly explanation.
Different from these efforts, this paper enables direct training and evaluation of consistent and verifiable language inference by providing a dataset that makes explicit the underlying evidence chains behind a high-level text classification task. 
\section{Conclusion and Discussion}
In this work, we proposed \texttt{TRIP}, a tiered benchmark dataset for physical commonsense reasoning posing a new challenge of jointly solving low-level to high-level tasks to form a coherent reasoning process. We experimented with several variations of tiered systems to solve the tasks. Our results show that in many cases, \textit{supervising large LMs based on high-level classification tasks in order to learn commonsense language understanding leads to inconsistent and unverifiable reasoning}, and inability to capture intermediate evidence toward the end task. 
Instead, we should train systems to jointly incorporate multiple types of lower-level evidence to solve reasoning tasks coherently.

Our detailed analysis of results offers strong intuition for future progress toward this goal.
As such, \texttt{TRIP} and our baselines provide an important first step toward verifiable, human-aligned commonsense language understanding, and a direction for development of AI systems in this area.\footnote{Our source code and data are publicly available at \url{https://github.com/sled-group/Verifiable-Coherent-NLU}.}

\paragraph{Broader impact.}
We use physical commonsense reasoning as an example in this work, but expect that a similar approach can apply to many aspects of language understanding. 
Our results have shown that a new challenge for the future will be to build machines that can reason logically and coherently, similar to what we expect from human reasoning. 
As these machines ultimately will work with humans, such alignment in reasoning is critical, as it will improve accountability and transparency in human-machine enterprise.

\section*{Acknowledgements}
This work was supported in part by the National Science Foundation (IIS-1617682 and IIS-1949634) 
and the DARPA
XAI program through UCLA (N66001-17-2-4029). We thank Bri Epstein and Haoyi Qiu for their assistance in annotation and hyperparameter tuning during this work, and the anonymous reviewers for their helpful comments and suggestions.

\bibliography{references_long}
\bibliographystyle{acl_natbib}

\clearpage

\appendix

\section{Physical State Annotations}\label{sec:datacollectionapx}
To collect our physical state annotations, we defined a space of 20 physical attributes (5 for humans, 15 for objects) which capture most conflicts found in the stories, collected in part from related attribute spaces proposed by \citet{gaoPhysicalCausalityAction2016} and \citet{bosselutSimulatingActionDynamics2018}. For humans, we track \textit{location}, \textit{hygiene}, and whether a human is \textit{conscious}, \textit{dressed}, or \textit{wet}. For objects, we consider \textit{location} and whether or not an object \textit{exists}, is \textit{clean}, connected to \textit{power}, \textit{functional}, \textit{in pieces}, \textit{wet}, \textit{open}, \textit{hot}, \textit{solid}, \textit{occupied} (i.e., containing another object), \textit{running} (i.e., turned on), \textit{movable}, \textit{mixed}, or \textit{edible}.

The values of these attributes each represent directions of physical state change (e.g., attribute became true or attribute became false), as listed in Appendix~\ref{sec:annotation1}. In the training data, we manually labeled each entity in the sentence with these attributes and values. For the other partitions, we used a semi-automatic approach described in Appendix~\ref{sec:annotation2}.

\subsection{Physical Annotation Label Space}\label{sec:annotation1}
When labeling entities for directions of physical state changes in sentences, we adopted the label space in Table~\ref{tab:condensed attribute space}. For predicting precondition and effect in non-location attributes as done in this work, it is straightforward to collapse this space into \textit{true}, \textit{false}, or \textit{unknown} for each. For human location labels, we use the full label space for predicting both precondition and effects for simplicity. Meanwhile, for object location labels, we simplify the problem by mapping them to smaller precondition and effect label spaces. While this does not significantly affect verifiability, this should be expanded in a full solution for better interpretability. For more detailed explanations, future work may consider tracking spans of text describing entity locations along the lines of \citet{amini2020procedural}.

\begin{table}[t]
    \centering\small
    \begin{tabular}{P{0.8cm}|P{1.7cm}|P{1.7cm}|P{1.7cm}}\toprule
         \textbf{Label} & \textbf{\thead{Human\\Location}} & \textbf{\thead{Object\\Location}} & \textbf{\thead{Other\\Attributes}}\\\midrule
         0 & irrelevant & irrelevant & irrelevant  \\\midrule
         1 & disappeared & disappeared & \textit{false} $\rightarrow$ \textit{false}  \\\midrule
         2 & moved & picked up & \textit{true} $\rightarrow$ \textit{true} \\\midrule
         3 & -- & put down & \textit{true} $\rightarrow$ \textit{false} \\\midrule
         4 & -- & put on & \textit{false} $\rightarrow$ \textit{true} \\\midrule
         5 & -- & removed & \_\_\_ $\rightarrow$ \textit{no} \\\midrule
         6 & -- & put in container & \_\_\_ $\rightarrow$ \textit{true} \\\midrule
         7 & -- & taken out of container & \textit{false} $\rightarrow$ \_\_\_ \\\midrule
         8 & -- & moved & \textit{true} $\rightarrow$ \_\_\_ 
         \normalsize \\\bottomrule
         
    \end{tabular}
    \caption{Label space and meanings for human location, object location, and other attributes. Each label represents a specific physical change (or lack of change). }
    \label{tab:condensed attribute space}
\end{table}

\subsection{Completing Physical State Annotations}\label{sec:annotation2}
\begin{figure}
    \centering
    \includegraphics[width=0.49\textwidth]{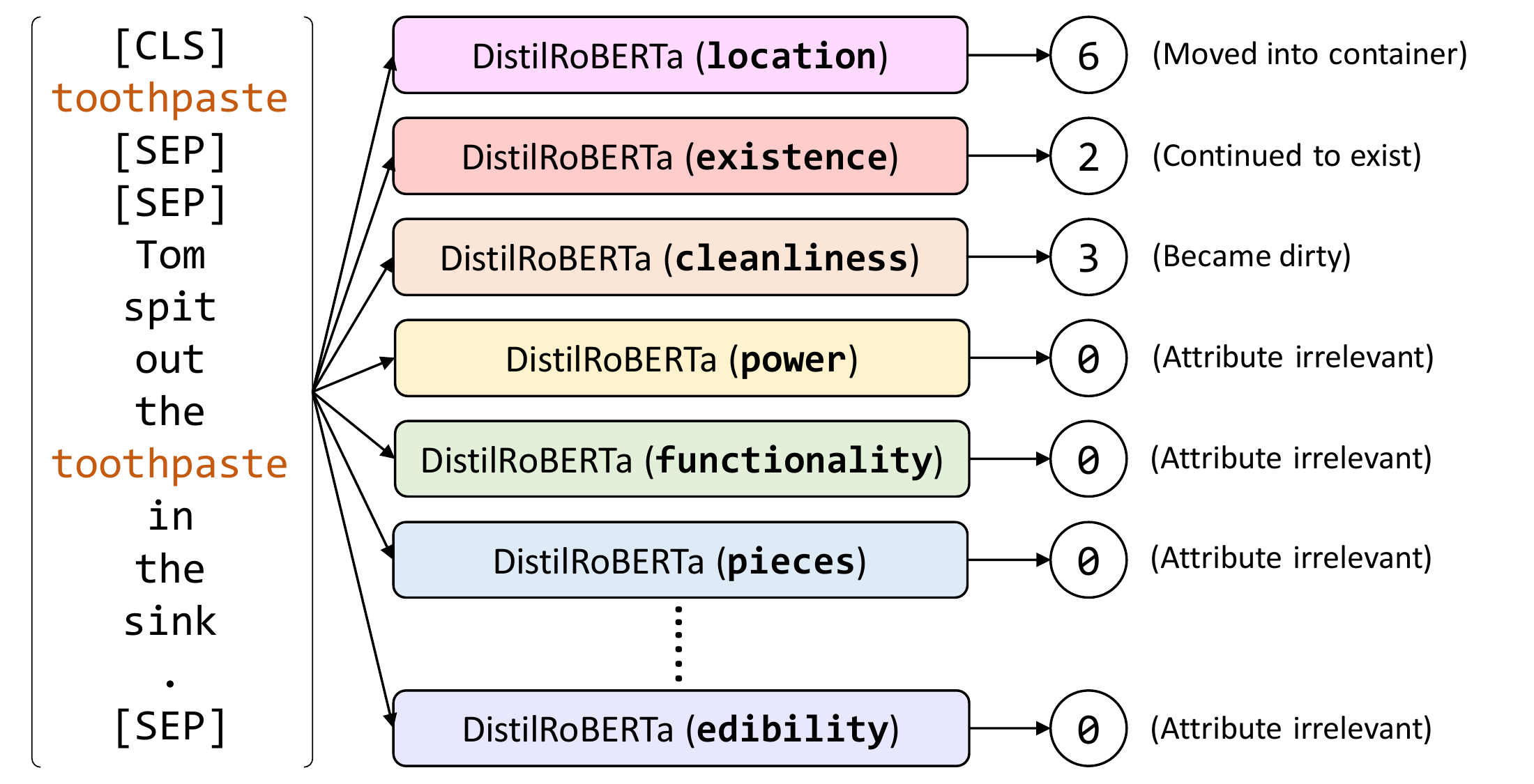}
    \caption{Proposed structure of the physical state classifier, consisting of 20 parallel instances of \textsc{DistilRoBERTa}. Each instance outputs an integer representing a particular kind of change (or lack of change) in the corresponding attribute.}
    \label{fig:state classifier}
\end{figure}

To expand our manual physical state annotations to the validation and testing data, we used the existing annotations to train classifiers to predict values for each attribute given a sentence-noun pair. First, each story was broken down into all possible sentence-noun pairs, using spaCy\footnote{\url{https://spacy.io/}} to identify noun phrases. These sentence-noun pairs were passed into the physical state classifier,\footnote{Followed \citet{guptaEffectiveUseTransformer2019} for formatting the input in order to generate entity-centric embeddings.} implemented as 20 parallel branches of \textsc{RoBERTa}, one for each physical attribute, as shown in Figure~\ref{fig:state classifier}. For efficiency, we use the pre-trained $\textsc{DistilRoBERTa}_{\textsc{BASE}}$ parameters (82M), distilled from $\textsc{RoBERTa}_{\textsc{BASE}}$ by \citet{liuRoBERTaRobustlyOptimized2019} with a small performance reduction~\cite{sanhDistilBERTDistilledVersion2019}. 
Using this module, we generated candidate physical state annotations for the remaining data, then manually revised them. As a different annotator completed this work from the annotator who completed the training data, we measured inter-annotator agreement on a representative subset of 157 sentences from 31 stories in the training data, finding a substantial Cohen's $\kappa$~\cite{cohenCoefficientAgreementNominal1960} of 0.7917. %To avoid divergence of the distributions of these annotations between partitions, we shuffled the data across partitions at this point.

\section{Model Implementation Details}
Each module in our tiered systems is implemented as some kind of neural network architecture. Here, we describe low-level details of the implementations.

\paragraph{Contextual Embedding.}
The Contextual Embedding module is implemented as a pre-trained transformer language model. Generally, this module takes as input a sentence and the name of an entity from a story, and outputs a dense numerical representation. We follow \citet{guptaEffectiveUseTransformer2019} in using an entity-first input to the language model to generate entity-centric embeddings. While there are some model-specific variations in special tokens, given an entity $e$ and a sentence $t_1, t_2, \cdots , t_n$, we structure the input sequence as `` $ \lbrack \text{CLS} \rbrack $ e $ \lbrack \text{SEP} \rbrack $ $t_1 \: t_2 \: \cdots \: t_n$  $ \lbrack \text{SEP} \rbrack $,'' where  $ \lbrack \text{CLS} \rbrack $ is a special token meant for input to classification layers, and $ \lbrack \text{SEP} \rbrack $ is a special separator token for multi-text inputs.

\paragraph{Precondition and Effect Classifiers.}
The Precondition and Effect Classifiers are implemented like typical classification heads for contextual embeddings, with one precondition classifier and one effect classifier for each of the 20 physical attribute tracked in the dataset. Specifically, each classifier is made up of two feedforward layers, each preceded by a dropout layer (using model specific defaults for dropout probability), with $\tanh$ activation in between them. The first layer performs a linear transformation on an input contextual embedding, while the second layer projects the hidden state to the size of the label space for the corresponding attribute. Argmax is applied to the output for classification. Altogether, the predictions from these classifiers label physical states of each entity in each sentence of the story.

\paragraph{Conflict Detector.}
For each entity and its predicted physical states over all sentences in a story, the Conflict Detector predicts whether there is some conflict in the entity's physical states, specifically flagging a pair of conflicting sentences through multi-label classification. Again, we use a transformer (6 additional layers with 8 attention heads) for this module, but model the high-level sequence of sentences in a story rather than the low-level sequence of tokens in a sentence.
For each sentence-entity pair, we consider the contextual embedding generated earlier, as well as the logits for all predicted precondition and effect states. We project both representations through linear layers to the same size, then concatenate them to form an entity dynamics representation. 
This representation for each sentence is input to the transformer, and the resulting hidden states are concatenated. Lastly, we use a feedforward layer followed by sigmoid activation to transform the hidden state to a belief probability of each sentence conflicting with another sentence in the story.

\paragraph{Story choice prediction.}
Given the output from the Conflict Detector, we lastly need to select which of the two given stories is plausible. As each Conflict Detector output represents the belief that a particular sentence conflicts with another sentence, we can simply sum the negative outputs for each story and apply softmax to determine which story is least likely to have a conflict.

\paragraph{Loss function details.}
To jointly train these various modules, we must balance several loss functions. 
The loss functions are weighted by corresponding scalar weights $\lambda_p$, $\lambda_f$, $\lambda_c$, and $\lambda_s$. In preliminary experiments, we found the best balance between state classification and the other tasks with the following assignment of weights: $\lambda_p = \lambda_f = \frac{0.4}{|A|}$, $\lambda_c = \lambda_s = 0.1$, where $|A|$ is the number of attributes tracked, i.e., 20. When omitting different loss functions, we rebalance the weights by ensuring $\lambda_c + \lambda_s = 0.2$, or $\lambda_c = \lambda_s$ where state classification losses are omitted.

\section{Model Training Details}\label{apx: training details}
The \textsc{RoBERTa}, \textsc{BERT}, and \textsc{DeBERTa} models are built from HuggingFace's \texttt{Transformers} library \cite{Wolf2019HuggingFacesTS}, particularly their implementation for multiple-choice classification, and the pre-trained $\textsc{BERT}_{\textsc{LARGE}}$ parameters (336M), $\textsc{RoBERTa}_{\textsc{LARGE}}$ parameters (355M), and $\textsc{DeBERTa}_{\textsc{BASE}}$ parameters (140M) respectively.
For all models, we use the AdamW optimizer~\cite{loshchilovDecoupledWeightDecay2018}.
Batch size is fixed at 1 story pair for all models, the maximum allowed by our GPU memory. To select the optimizer learning rate and number of training epochs, all models are trained by grid search over these two, maximizing the validation set verifiability as defined in Section~\ref{sec:metrics}. Learning rate is selected from the set $\{ 1 \times 10^{-6},  5 \times 10^{-6},  1 \times 10^{-5}, 5 \times 10^{-5}, 1 \times 10^{-4} \}$, while the maximum number of epochs is fixed at 10. Ties are broken first by validation accuracy on the end plausibility classification task, then by selecting the model instance trained for fewer epochs (to avoid overfitting).
The selected learning rate and number of epochs for each model presented in the main paper are listed in Table~\ref{tbl:hyperparam search}. 

\begin{table}[t]
    \centering
    \scriptsize
    
    \begin{tabular}{P{2.0cm}|P{2.0cm}P{2.0cm}}\toprule
           \textbf{Model} & \textbf{Learning Rate} & \textbf{Epochs}\\\midrule
        \multicolumn{3}{c}{\textit{Table~\ref{tbl:loss results}, All Losses }}  \\\midrule
        \textsc{BERT} &  5e-6 & 5   \\
        \textsc{RoBERTa} & 1e-5 & 8   \\
        \textsc{DeBERTa} & 5e-6 & 6   \\\midrule
        \multicolumn{3}{c}{\textit{Table~\ref{tbl:loss results}, Omit Story Choice Loss }}  \\\midrule
        \textsc{BERT} & 5e-5 & 9   \\
        \textsc{RoBERTa}  & 1e-5 & 6 \\
        \textsc{DeBERTa}  & 5e-5 & 8  \\\midrule    
        \multicolumn{3}{c}{\textit{Table~\ref{tbl:loss results}, Omit Conflict Detection Loss }}  \\\midrule
        \textsc{BERT} & 1e-6 & 2    \\
        \textsc{RoBERTa}  & 5e-6 & 9  \\
        \textsc{DeBERTa}  & 1e-6 & 4  \\\midrule         
        \multicolumn{3}{c}{\textit{Table~\ref{tbl:loss results}, Omit State Classification Loss }}  \\\midrule
        \textsc{BERT} & 1e-5 & 4     \\
        \textsc{RoBERTa}  & 1e-6 & 8  \\
        \textsc{DeBERTa}  & 5e-6 & 10   \\

        \bottomrule          
    \end{tabular}

    \normalsize
    \caption{Selected learning rate (LR), number of training epochs, and validation verifiability and accuracy for all results presented in the paper.}
    \vspace{-5pt}
    \label{tbl:hyperparam search}
\end{table}

\section{Supplementary Results}\label{sec:extraresults}
Lastly, we provide additional results that were omitted from the main paper.\footnote{Note that the results in this appendix use a slightly simpler label space for \texttt{location} state classification, and thus are not directly comparable to the results presented in the main paper.}

\subsection{Conflict Detector Ablations}\label{sec:ablation results}
The Conflict Detector module takes in two types of inputs: 1) contextual embeddings of sentence-entity pairs, and 2) physical state logits from the Precondition and Effect Classifiers. To determine the impact of each, we present ablations omitting them for the best-performing instances from the previous section, i.e., those not considering story choice classification loss. Table~\ref{tbl: ablations validation} presents these results for the validation set, while Table~\ref{tbl: ablations} presents these results for the test set.

Without including the physical state inputs, we see a slight drop in consistency and verifiability of some models. For example, \textsc{RoBERTa} drops from 9.7\% verifiability and 23.4\% consistency to 4.6\% and 17.7\%, respectively. Meanwhile, \textsc{DeBERTa} increases from 8.0\% verifiabiliy and 20.2\% consistency to 11.4\% and 24.5\%. While \textsc{RoBERTa} seems to depend slightly on the predicted physical states in performing conflict detection, \textsc{DeBERTa} favors the contextual embedding.

Without including the contextual embeddings, we see a drastic drop across the board to below-random performance, with \textsc{RoBERTa} dropping to 0\% verifiability and consistency, and \textsc{DeBERTa} to 2.3\% and 6.6\% respectively. This suggests that while forcing the model to track physical states enables greater explanation, they are not sufficient for models to learn conflict detection, or they are not
incorporated successfully into the higher-level predictions. The contextual embedding, which is fine-tuned on physical state classification and conflict detection jointly, seems to be most powerful for solving the end task.
Future work should further explore how to harness the rich information provided by the physical states to improve system performance and interpretability.

% *********** validation results **************
\begin{table}[t]
    \centering
    \scriptsize
    
    \begin{tabular}{P{1.1cm}|P{0.5cm}P{0.5cm}|P{1.0cm}P{0.8cm}P{1.1cm}}\toprule
            & \textbf{Verif.} & \textbf{Acc.}  & \textbf{Prec. F1} & \textbf{Eff. F1}  & \textbf{Confl. F1} \\
            \textbf{Model} &  (\%) & (\%)  &   (\%) & (\%) & (\%)   \\\midrule
        \multicolumn{6}{c}{\textit{Contextual Embeddings + Physical States}}  \\\midrule            
        \textsc{BERT} & 9.6 & 70.2 & 74.4 & 66.7 & 65.1    \\
       \textsc{RoBERTa}  & \textbf{12.1} & \textbf{77.0} & 72.3 & 62.7 & 70.9 \\
        \textsc{DeBERTa}  & 11.2 & 72.7 & 77.0 & 71.1 & 68.2   \\\midrule            
        \multicolumn{6}{c}{\textit{Contextual Embeddings Only}}  \\\midrule
        \textsc{BERT} &  \textbf{10.9} & 72.7 & 75.9 & 69.3 & 66.7    \\
        \textsc{RoBERTa} & 9.6 & \textbf{76.1} & 72.5 & 61.6 & 70.3 \\
        \textsc{DeBERTa} & 9.9 & \textbf{76.1} &  77.3 & 71.3 & 68.6   \\\midrule
        \multicolumn{6}{c}{\textit{Physical States Only}}  \\\midrule
        \textsc{BERT} & 0.6 & 54.7 & 60.5 & 59.9 & 51.1    \\
        \textsc{RoBERTa}  & 0.0 & 43.2 & 38.4 & 37.8 & 49.5 \\
        \textsc{DeBERTa}  & 2.2 & 58.1 & 81.0 & 79.0 & 53.0   \\\bottomrule      
    \end{tabular}

    \normalsize
    \caption{Validation set performance of best models in Table~\ref{tbl:loss results} when ablating inputs to the Conflict Detector.}
    \vspace{-5pt}
    \label{tbl: ablations validation}
\end{table}

\begin{table}[t]
    \centering
    \footnotesize
    
    \begin{tabular}{P{1.4cm}|P{1.3cm}P{1.6cm}P{1.6cm}}\toprule
            & \textbf{Accuracy} & \textbf{Consistency} & \textbf{Verifiability} \\
            \textbf{Model} &  (\%) & (\%)  &   (\%)    \\\midrule
        % {random} &  49.5$\pm$0.2 &  10.7$\pm$0.3  & 0.0$\pm$0.0   \\\midrule
        % {random} &  49.5 &  10.7 & 0.0 \\\midrule 
          
        \multicolumn{4}{c}{\textit{Contextual Embeddings + Physical States}}  \\\midrule
        % \textsc{BERT} & 7.4 & 15.7 & 63.2    \\
        \textsc{BERT} & 63.2  & 15.7 & 7.4  \\
        % \textsc{RoBERTa}  & 9.7 & 23.4 & \textbf{76.6}  \\
        \textsc{RoBERTa}  &  \textbf{76.6} & 23.4 & 9.7  \\
        % \textsc{DeBERTa}  & 8.0 & 20.2 & 72.9    \\\midrule    
        \textsc{DeBERTa}  & 72.9  & 20.2 & 8.0    \\\midrule    
        \multicolumn{4}{c}{\textit{Contextual Embeddings Only}}  \\\midrule
        % \textsc{BERT} &  6.8 & 16.8 & 70.7     \\
        \textsc{BERT} & 70.7 & 16.8 & 6.8    \\
        % \textsc{RoBERTa} & 4.6 & 17.7 &  \textbf{76.6}  \\
        \textsc{RoBERTa} & \textbf{76.6} & 17.7 &  4.6    \\
        % \textsc{DeBERTa} & \textbf{11.4} & \textbf{24.5 }&  74.1   \\\midrule
        \textsc{DeBERTa} & 74.1  & \textbf{24.5} & \textbf{11.4}    \\\midrule
        \multicolumn{4}{c}{\textit{Physical States Only}}  \\\midrule
        % \textsc{BERT} & 0.3 & 3.4 &  56.1     \\
        \textsc{BERT} &  56.1 & 3.4 &   0.3    \\
        % \textsc{RoBERTa}  & 0.0 & 0.0  & 42.2  \\
        \textsc{RoBERTa}  &  42.2  & 0.0  & 0.0 \\
        % \textsc{DeBERTa}  & 2.3 & 6.6 &  59.3    \\\bottomrule      
        \textsc{DeBERTa}  &  59.3 & 6.6 & 2.3     \\\bottomrule   
    \end{tabular}

    \normalsize
    \caption{Validation set performance of best models in Table~\ref{tbl:loss results} when ablating inputs to the Conflict Detector.}
    \vspace{-5pt}
    \label{tbl: ablations}
\end{table}

\subsection{State Classification Performance by Attribute}
\begin{figure}
    \centering
    \includegraphics[width=1.0\linewidth]{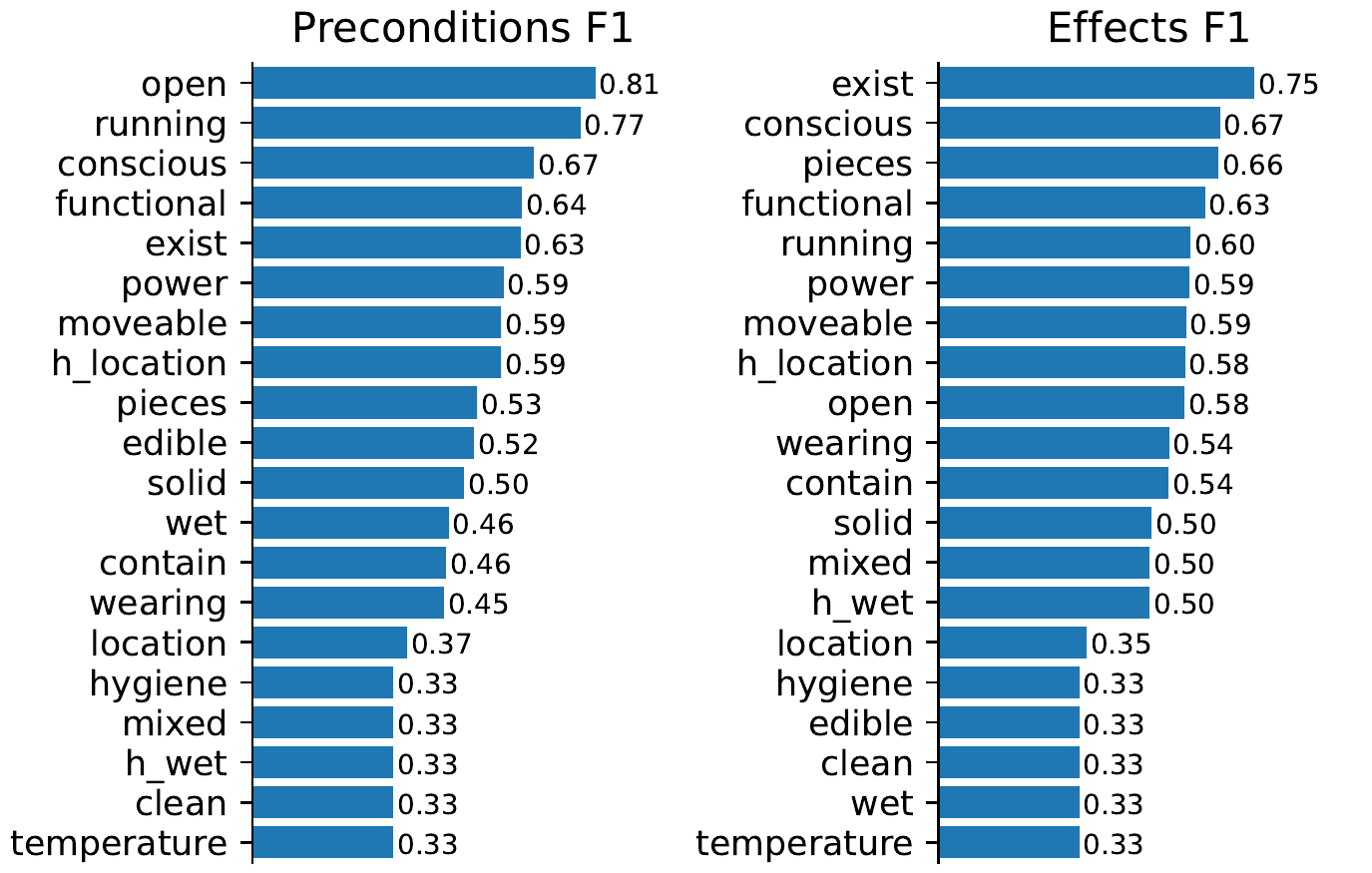}
    \caption{Precision and recall of predictions for each attribute from our best RoBERTa model on the validation set. }
    \label{fig:error_analysis_phys}
\end{figure}

Figure~\ref{fig:error_analysis_phys} breaks down the F1 score for predicting precondition and effect states by attribute across the \texttt{TRIP} dataset. We find that for preconditions, openness and whether objects are running, i.e., activated, are best captured, and for effects, existence and consciousness are. Meanwhile, wetness and temperature are challenging for predicting both preconditions and effects.

\end{document}